\documentclass[11pt,a4paper]{article}
\usepackage{acl}
\usepackage{times}
\usepackage{latexsym}
\usepackage{booktabs}
\usepackage{amsmath,amssymb}
\usepackage{graphicx}
\usepackage{multirow}
\usepackage{subcaption}
\usepackage{tcolorbox}
\usepackage{cleveref}
\usepackage{lipsum}
\usepackage{xcolor,colortbl}
\usepackage[T1]{fontenc}
\newcommand{\colorSquare}[1]{\raisebox{0.7mm}{\colorbox{#1}{\rule{0pt}{0.8mm}\rule{1mm}{0pt}}}}
\definecolor{colorD}{HTML}{006D77} 
\definecolor{colorC}{HTML}{83C5BE} 
\definecolor{colorB}{HTML}{FFDDD2} 
\definecolor{colorA}{HTML}{E29578}

\usepackage{microtype}

\title{DateLogicQA: Benchmarking Temporal Biases in Large Language Models}

\author{Gagan Bhatia~$^{\phi}$, MingZe Tang~$^{\phi}$, Cristina Mahanta~$^{\phi}$, Madiha Kazi~$^{\phi}$ \\
  University of Aberdeen \\
\texttt{\{g.bhatia.24,m.tang.24,c.mahanta.24,m.kazi.24\}@abdn.ac.uk}
}

\date{}

\begin{document}
\maketitle
\begin{abstract}

We introduce \textbf{DateLogicQA}, a human curated benchmark of \textit{190} questions specifically designed to understand temporal bias in Large Language Models (LLMs). Covering seven date formats across past, present, and future contexts, DateLogicQA examines four reasoning types: commonsense, factual, conceptual, and numerical. Through human-led evaluations of 12 state-of-the-art LLMs, we identify Representation-Level Bias, arising from suboptimal embeddings that distort date semantics, and Logical-Level Bias, manifesting when correct date tokens yield flawed temporal reasoning. Our findings underscore persistent challenges in handling various date formats and temporal contexts, revealing the need for more robust pretraining data, targeted post-training methods, and precise tokenization strategies. By illuminating these biases, we provide actionable insights to guide the development of LLMs for accurate temporal reasoning across diverse real-world applications.

\end{abstract}

\section{Introduction}

Accurate temporal reasoning is essential for real-world applications like event planning and historical questions. However, biases in Large Language Models (LLMs) can lead to misinterpretations or errors in date-related tasks. Understanding these biases is essential for precisely handling numerical structures and contextual meanings, making temporal reasoning ideal for identifying and analysing biases in tokenization, representation, and logical reasoning.

A significant source of these biases originates from the tokenization process. While tokenizers divide the text into subword units, inconsistencies in tokenizing dates can disrupt reasoning tasks. This can lead to two types of biases: Representation-Level Bias, caused by inconsistencies in embeddings affecting semantic structures of dates, and Logical-Level Bias, where correct tokens do not yield accurate outputs due to misaligned internal processing. Together, these biases highlight the challenges LLMs face in preserving the integrity and interpretability of temporal data across diverse formats and contexts.

\begin{figure}[t]
\includegraphics[width=\columnwidth]{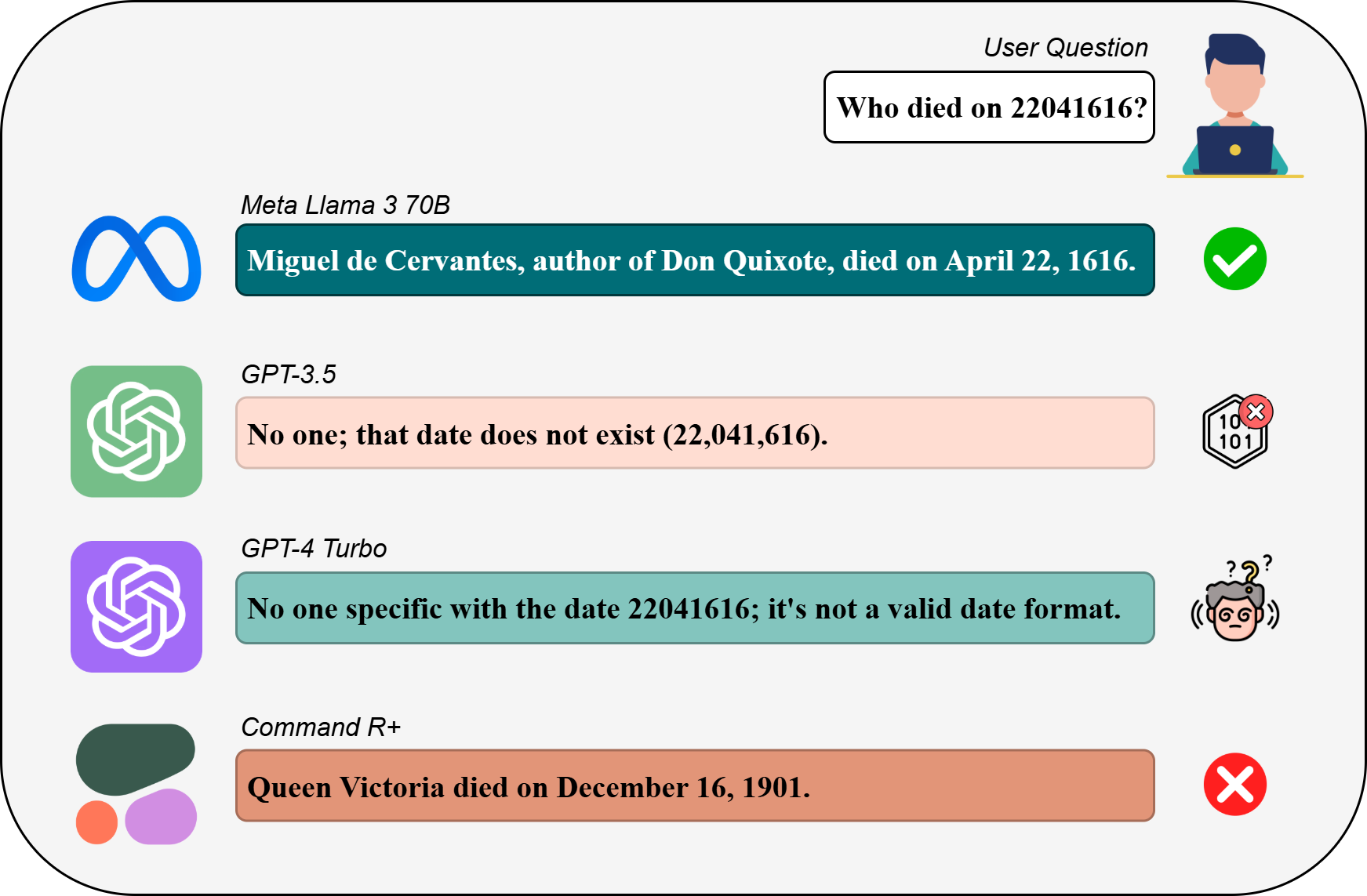}
\centering
\caption{Examples of temporal biases in LLMs. \colorSquare{colorA} \textbf{Incorrect} Response,\colorSquare{colorB} \textbf{Faulty Date} but accurate reasoning indicating representation level temporal bias, \colorSquare{colorC} \textbf{Faulty reasoning} but accurate date indicating logical level temporal bias, \colorSquare{colorD} \textbf{Correct} response }
\label{fig:intro}
\end{figure}

This paper makes two significant contributions to understanding temporal biases in LLMs. \textbf{(1)} We introduce \textbf{DateLogicQA}, a dataset of $190$ curated questions for evaluating temporal reasoning across various date formats, contexts (past, present, future), and reasoning types (commonsense, factual, conceptual, numerical). 
\textbf{(2)} We conduct human evaluations of model responses to analyse tokenization accuracy and reasoning quality, providing insights beyond automated metrics.

We have organised the paper as follows: Section \ref{sec:related_works} reviews related works, summarising the impact of tokenization on LLM performance and past temporal reasoning approaches. Section \ref{sec:dataset} details the creation of the DateLogicQA dataset, including its design principles and examples. Section \ref{sec:Methodology} outlines methods for temporal reasoning, and biases. Section \ref{sec:Results} presents experiment results, followed by a discussion of findings and bias mitigation in Section \ref{sec:discussion}. Lastly, Section \ref{sec:conclusion} summarises our contributions.

\section{Related Works} \label{sec:related_works}

\paragraph{Impact of Tokenization on Language Models}

Tokenization significantly affects the efficiency and reasoning abilities of large language models (LLMs). Research by \citet{gu2024retok} and \citet{goldman2024unpacking} highlights that tokenizers with higher compression rates enhance representation efficiency, particularly in smaller models. However, \citet{schmidt2024tokenization} argue that effective tokenization also depends on pre-tokenization and vocabulary design. Studies like \citet{ahia2023do} show that poorly tokenized languages face performance and fairness issues. Furthermore, choices in tokenization impact reasoning; \citet{zhang2024counting} and \citet{singh2024tokenization} indicate that numerical tokenization can lead to errors in arithmetic and counting tasks. \citet{rajaraman2024toward}, \citet{alberts2024interleaving}, \citet{minixhofer2024zero-shot}, and \citet{gastaldi2024the} show how well-designed tokenizers improve sequence pattern modelling and numerical reasoning through advanced embedding methods. Our study extends this work by examining tokenization's role in handling diverse date formats for temporal reasoning.

\paragraph{Temporal Reasoning in LLMs}

Temporal reasoning poses challenges for LLMs due to inherent biases. \citet{zhu2024is} discussed "nostalgia bias" (favouring outdated knowledge) and "neophilia bias" (speculative future predictions), while \citet{tan2023towards} observed inconsistent generalisation across different time periods. Structured approaches like temporal graphs \cite{xiong2024large} and synthetic datasets \cite{fatemi2024test} enhance performance by explicitly encoding temporal relationships. Additionally, tokenization critically affects temporal reasoning; \citet{zhao2024set} found that temporal misalignment hampers accuracy, and \citet{kishore2024unveiling} identified inductive biases in models like GPT-3.5 and GPT-4. \citet{su2024timo} propose task-agnostic approaches to enhance temporal reasoning, while \citet{gastaldi2024the} and \citet{rajaraman2024toward} link tokenization to reasoning performance. By analysing how tokenization strategies affect temporal reasoning, especially for date formats, our work fills a gap in understanding the interplay between tokenization and temporal task performance.

\section{DateLogicQA} \label{sec:dataset}

We introduce \textbf{DateLogicQA}, a dataset designed to explore how LLMs handle dates in various formats and contexts to tokenize, interpret, and reason with them. It consists of 190 questions divided into four categories: \textit{commonsense}, \textit{factual}, \textit{conceptual}, and \textit{numerical}. Each category features one of seven date formats across three temporal contexts: \textit{past}, \textit{present}, and \textit{future}. This systematic variation allows for an in-depth analysis of LLMs' performance with temporal information.

\paragraph{Objective and Purpose}
The dataset aims to assess LLMs' tokenization and understanding of dates, as errors can lead to interpretative biases. By embedding dates within questions, we evaluate context-rich date interpretation, simulate real-world scenarios where dates carry contextual significance, and test LLMs' ability to extract and interpret date information accurately.

\begin{table}[!htp]\centering
\scriptsize
\resizebox{\columnwidth}{!}{%
\begin{tabular}{ll}
\toprule
\textbf{Concepts} & \textbf{Example} \\ \midrule
\multirow{2}{*}{\textbf{Numerical}} & 
What is the time 7 years and 9 months  \\ 
& after 27101446? \\\midrule
\multirow{3}{*}{\textbf{Factual}} & 
Which of the people died on 23041616? \\
& A) Shah Jahan B) Miguel de Cervantes \\
& C) Princess Diana D) William Shakespeare 
\\\midrule
\multirow{2}{*}{\textbf{Conceptual}} & 
The first iPhone was released on 29062007. \\
& How many years has it been since its release? \\\midrule
\multirow{3}{*}{\textbf{Commonsense}} & 
John was born on 15-03-1985. \\
& He graduated from college on 01-05-2007. \\ 
& Was John older than 18 when he graduated? \\
\bottomrule
\end{tabular}
}
\caption{Dataset samples illustrating different temporal reasoning concepts.}\label{tab: examples}
\end{table}

\begin{table}[!htp]
\centering
\resizebox{\columnwidth}{!}{%
\begin{tabular}{ll}
\toprule
\textbf{Date Format}          & \textbf{Example}  \\ \midrule
DDMMYYYY                      & 23041616          \\ 
MMDDYYYY                      & 04231616           \\ 
DDMonYYYY                     & 23April1616       \\ 
DD-MM-YY                      & 23-04-16          \\ 
YYYY, Mon DD                  & 1616, April 23    \\ 
DD/YYYY (Julian calendar)     & 113/1616          \\ 
YYYY/DD (Julian calendar)     & 1616/113          \\ 
\bottomrule
\end{tabular}}
\caption{Dataset samples illustrating different date formats used.}
\label{tab:date_formats}
\end{table}

This approach comprehensively examines various temporal notations, including uncommon formats like Julian calendar representations.

\paragraph{Temporal Distribution}
DateLogicQA spans a broad temporal range, featuring dates from historical periods (e.g., the 1600s), modern contexts (e.g., the 2000s), and hypothetical futures (e.g., the 2100s). For clarity, we categorised dates into \textit{past}, \textit{present}, and \textit{future}, with some questions covering multiple dates to assess LLMs' ability to manage temporal relationships across contexts.

\paragraph{Rationale for Design}
The dataset prioritises models' ability to interpret dates within broader narratives rather than as isolated data points. Its smaller size allows for careful curation of high-quality, linguistically diverse questions, focusing on specific nuances of temporal reasoning. This enables detailed analysis of model behaviour and understanding of temporal biases.
 \section{Methodology} \label{sec:Methodology}

 \subsection{Human-Led Temporal Bias Assessment}


Understanding temporal contexts is crucial for analysing events over time. This includes grasping temporal references like \textit{"How many years has it been since..."} (Past) and \textit{"What will the contract's last day be..."} (Future), along with the maintenance of logical chronological order and adaptation to changes in context. For large language models, this capability is vital for tasks such as historical inquiries, time-sensitive query handling and predictions about future events. Assessing biases in temporal reasoning is essential for accuracy across various applications. We utilized the dataset referenced in Section 3.

\begin{figure}[t]
\includegraphics[width=\columnwidth]{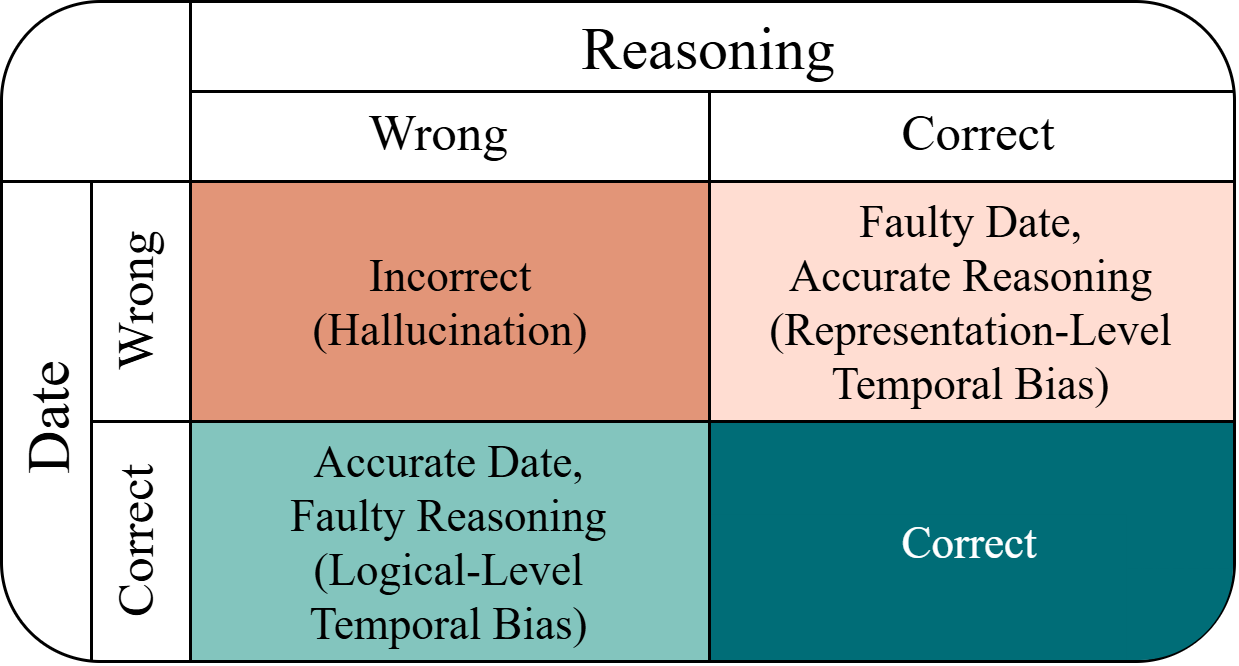}
\centering
\caption{Human evaluation rubric}
\label{fig:color_chartrq2}
\end{figure}

We conduct a human evaluation to assess the temporal bias of LLMs as automated methods may exhibit inherent biases that affect results, ultimately undermining the evaluation's purpose. This methodology provides a more reliable analysis, identifying outliers that respond accurately without fully comprehending temporal aspects. Instead, it relies on contextual clues or learned patterns acquired during training or through retrieval-augmented generation.

Model responses are categorised based on colours in \autoref{fig:color_chartrq2}, representing levels of temporal understanding. \textbf{Dark Orange} (\colorSquare{colorA}) denotes incorrect answers or temporal hallucinations from failure to tokenize dates or grasp context. \textbf{Light Orange} (\colorSquare{colorB}) reflects Representation-Level Temporal Bias, where the model tokenizes dates inaccurately but reaches the correct answer through logical reasoning. This suggests that some internal reasoning within the model compensates for misunderstanding the date format. \textbf{Light Teal} (\colorSquare{colorC}) signifies Logical-Level Temporal Bias, where the model tokenizes correctly but misapplies logic due to misattributing events or calculation errors. Finally, \textbf{Dark Teal} (\colorSquare{colorD}) denotes correct answers, indicating successful tokenization and logical reasoning. This illustrates a complete understanding of the question.

\section{Results} \label{sec:Results}


\begin{figure*}[h!]
    \centering
    \begin{subfigure}[t]{0.3\textwidth}
        \centering
        \includegraphics[width=\textwidth]{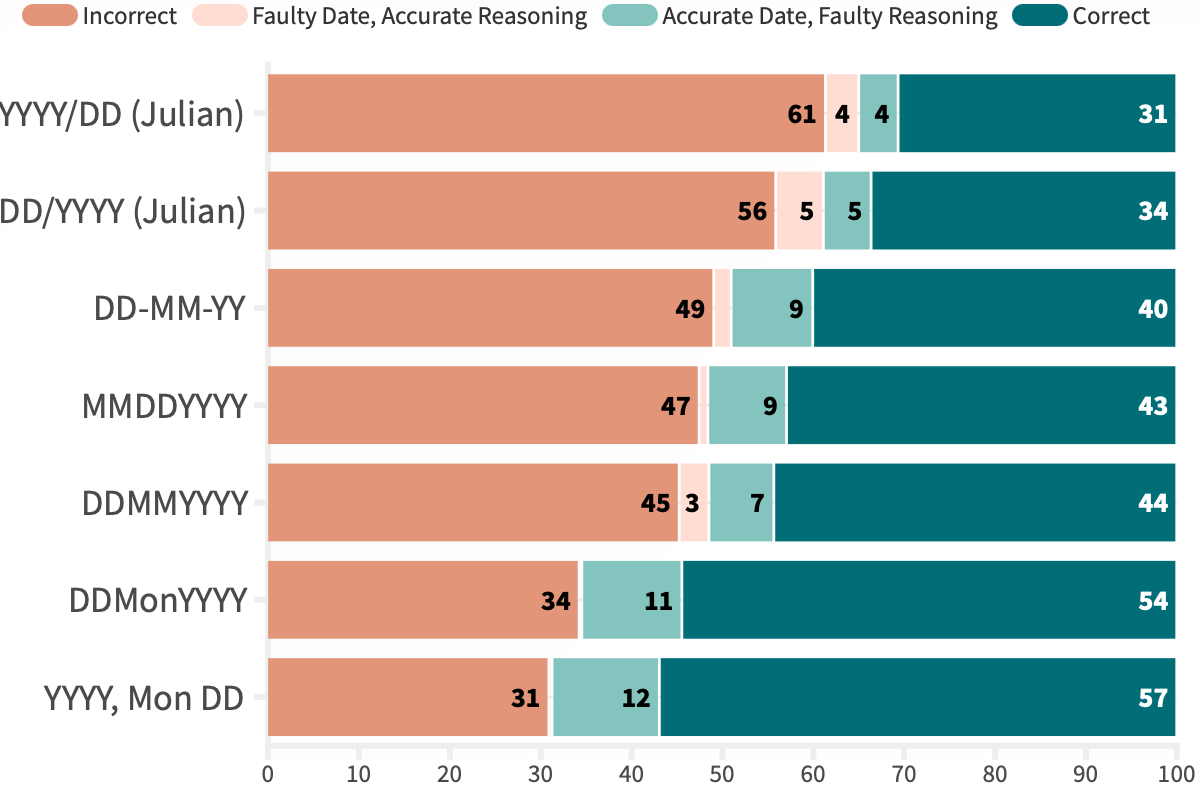}
        \caption{Date Format visualisation}
        \label{fig:embeddings}
    \end{subfigure}
    \begin{subfigure}[t]{0.3\textwidth}
        \centering
        \includegraphics[width=\textwidth]{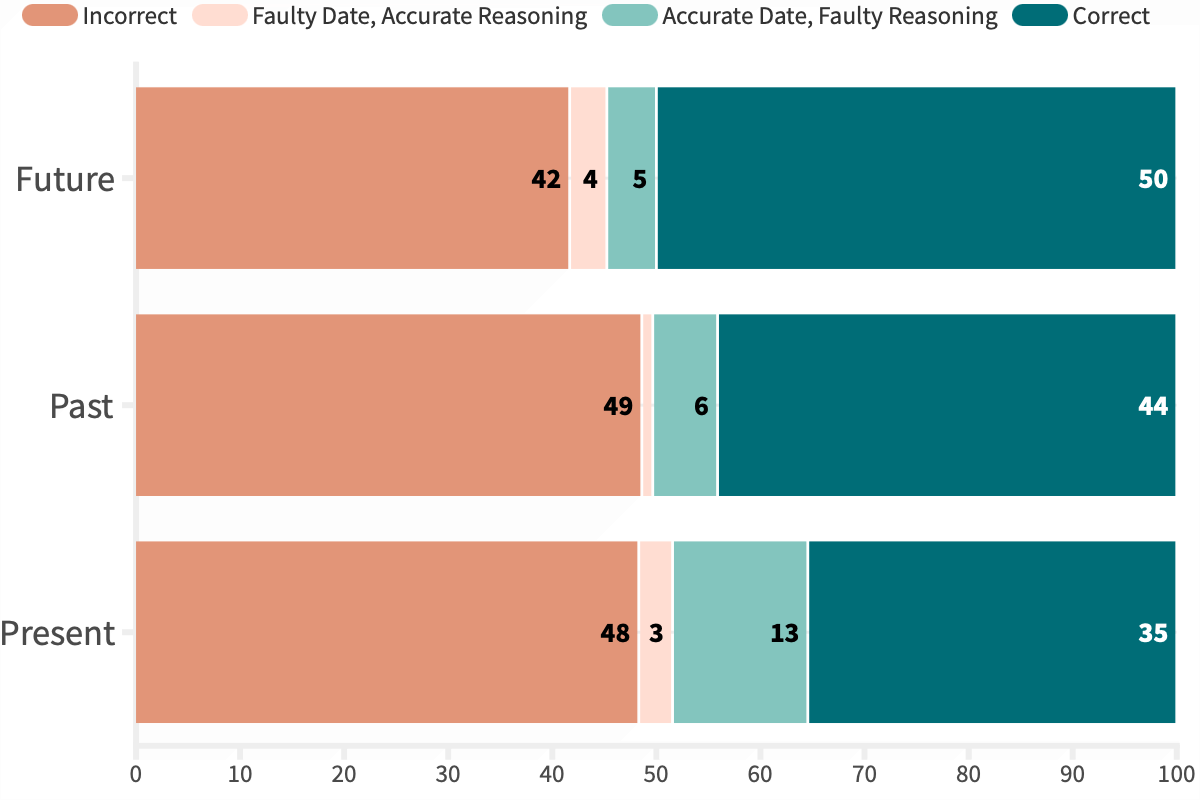}
        \caption{Time period visualisation}
        \label{fig:softmax}
    \end{subfigure}
    \begin{subfigure}[t]{0.3\textwidth}
        \centering
        \includegraphics[width=\textwidth]{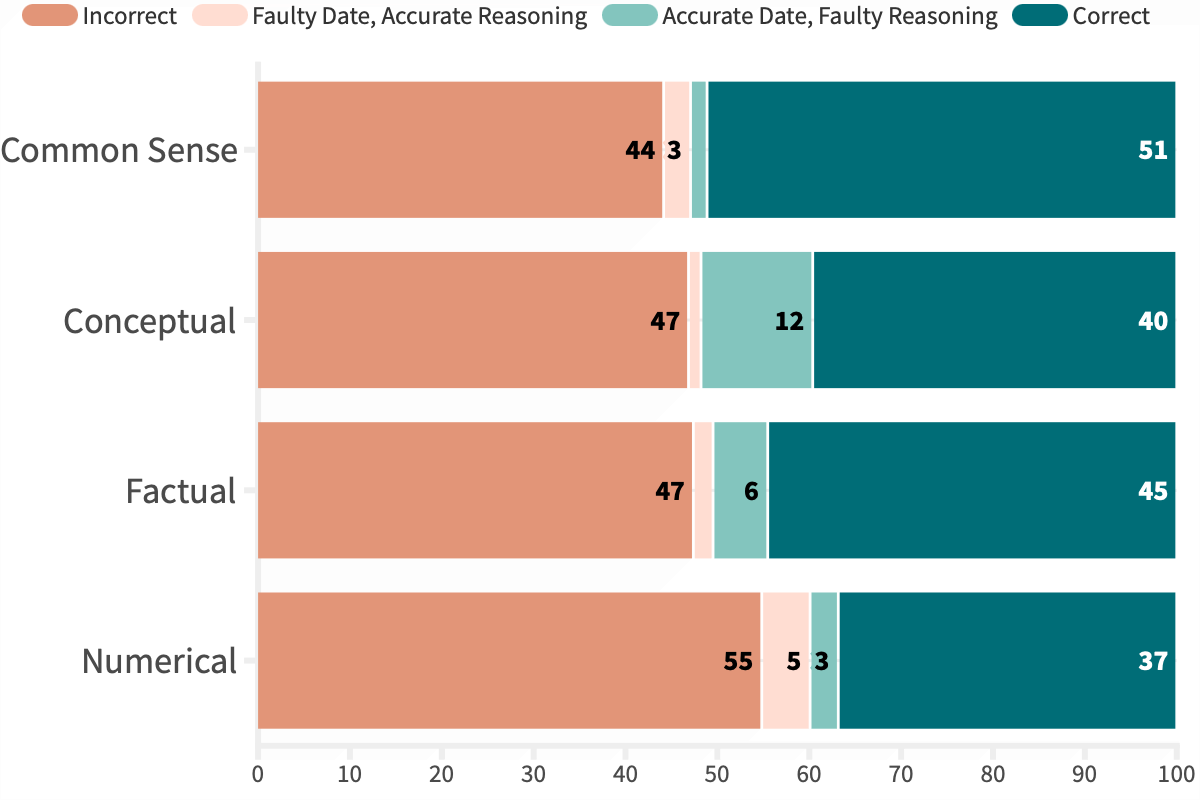}
        \caption{Question Type visualisation}
        \label{fig:softmax}
    \end{subfigure}
    \caption{Results Visualisations}
    \label{fig:}
\end{figure*}
\begin{figure}[t]
\centering
\includegraphics[width=1\columnwidth]{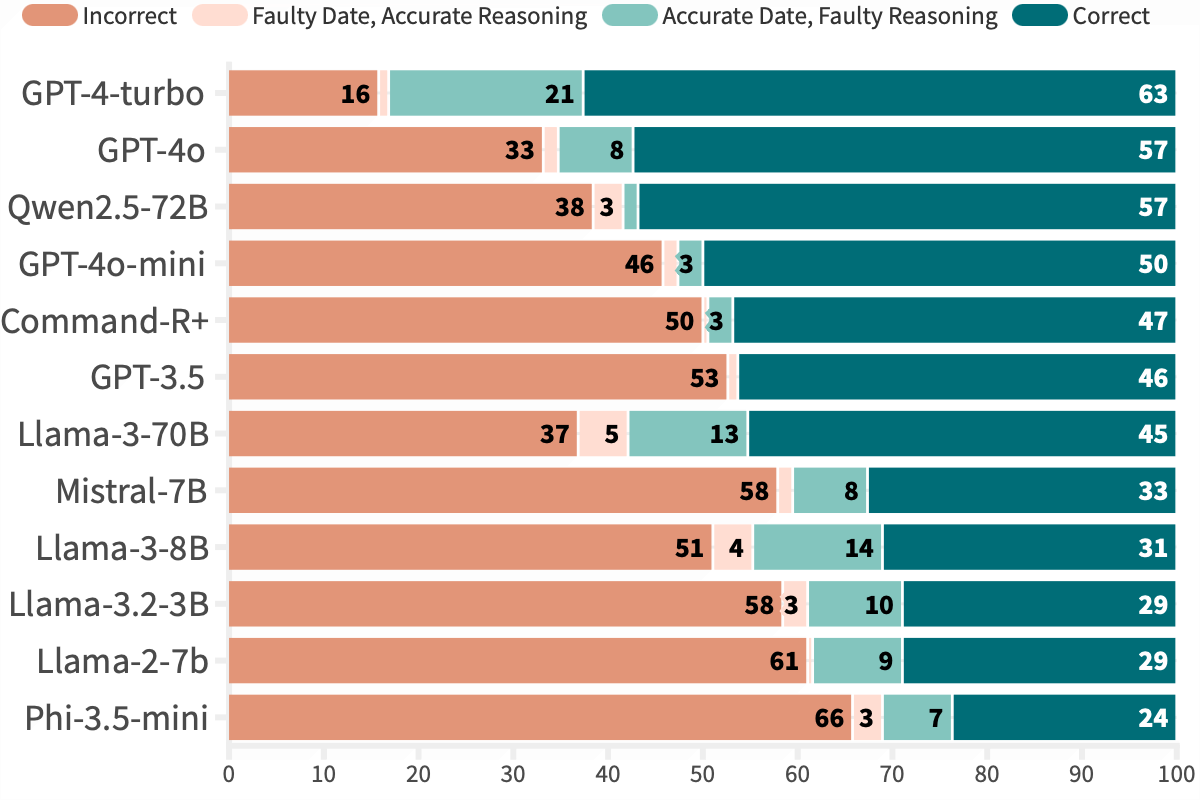}
\caption{Each bar is segmented into four colors representing the quality of responses: \colorSquare{colorA} \textbf{Incorrect} Response,\colorSquare{colorB} \textbf{Faulty Date} but accurate reasoning indicating representation level temporal bias, \colorSquare{colorC} \textbf{Faulty reasoning} but accurate date indicating logical level temporal bias, \colorSquare{colorD} \textbf{Correct} response}
\label{fig:ResultsRq2}
\end{figure}

\subsection{Temporal Reasoning Analysis}
\begin{table*}[!htp]
\centering
\scriptsize
\resizebox{\textwidth}{!}{%
\begin{tabular}{l p{0.1\textwidth} p{0.1\textwidth} l p{0.1\textwidth} p{0.1\textwidth} p{0.1\textwidth} p{0.1\textwidth} p{0.1\textwidth}}
\toprule
\textbf{Type of Q} & \textbf{Question} & \textbf{Answer} & \textbf{Date Format} & \textbf{Llama3-70B} & \textbf{Qwen2.5-72B} & \textbf{GPT-4o} & \textbf{GPT-4-turbo} & \textbf{CMDR+} \\
\midrule

Factual &
Which of the following famous people died on 23041616? Answer only with A, B, C, or D:
A) Shah Jahan 
B) Miguel de Cervantes 
C) Princess Diana 
D) William Shakespeare &
William Shakespeare \textbf{OR} Miguel de Cervantes &
DDMMYYYY &
\cellcolor[HTML]{006D77}\textcolor{white}{B) Miguel de Cervantes and D) William Shakespeare} &
\cellcolor[HTML]{006D77}\textcolor{white}{B, D} &
\cellcolor[HTML]{006D77}\textcolor{white}{B) Miguel de Cervantes} &
\cellcolor[HTML]{006D77}\textcolor{white}{D)William Shakespeare} &
\cellcolor[HTML]{E29578}A \\
\midrule

Conceptual &
The Eiffel Tower was built between 028/1887 to 090/1889. How long did it take to build the Eiffel Tower? &
2 years, 2 months, and 3 days &
DD/YYYY (Julian) &
\cellcolor[HTML]{E29578}The Eiffel Tower took 2 years and 1 month to build. &
\cellcolor[HTML]{E29578}2 years, 10 months, 2 days. &
\cellcolor[HTML]{E29578}2 years and 1 month. &
\cellcolor[HTML]{006D77}\textcolor{white}{It took 2 years and 2 months to build.} &
\cellcolor[HTML]{006D77}\textcolor{white}{Construction of the Eiffel Tower took about two years and two months.} \\
\midrule

Common Sense &
A new electric car model is set to be released on 273/2040. Pre-orders open 6 months earlier. On what date can customers start preordering? &
On March 30, 2040 &
DD/YYYY (Julian) &
\cellcolor[HTML]{E29578}273/2040 - 6 months = 93/2039. Customers can start preordering on this date. &
\cellcolor[HTML]{E29578}Pre-orders open on 273/2039. &
\cellcolor[HTML]{E29578}Customers can start pre-ordering on March 273/2039. &
\cellcolor[HTML]{E29578}Customers can start preordering on March 273, 2040. &
\cellcolor[HTML]{E29578}The pre-order date is July 273, 2039. \\
\midrule

Numerical &
What is the time 7 years and 9 months after 10271446? &
July 27, 1454 &
MMDDYYYY &
\cellcolor[HTML]{E29578}October 27, 2040. &
\cellcolor[HTML]{E29578}10271446 + 7 years 9 months = 10353406 &
\cellcolor[HTML]{E29578}October 2023 plus 7 years and 9 months is July 2031. &
\cellcolor[HTML]{E29578}Time: 10429846 (Unix timestamp format). &
\cellcolor[HTML]{E29578}10279141 \\
\bottomrule
\end{tabular}}
\caption{Model Performance on DateLogicQA}\label{tab:temporal_reasoning_compact}
\end{table*}

Temporal reasoning, including processing and drawing inferences from historical and future dates, is one of the most challenging tasks for large language models. The current study investigates whether there are any differences in LLM performance when reasoning with historical dates, such as "July 20, 1969", and future dates, such as "January 1, 2050". To this end, we present the testing of 12 state-of-the-art LLMs using a question-answer dataset encompassing different date formats and various temporal contexts. This paper examines their skills in tokenization, comprehension, and inference on dates. We classify the answers into four categories based on their accuracy and treatment of the dates and logical structure involved, thereby providing a systematic evaluation framework.

In order to ensure that the assessment is robust, four human annotators, each with at least four years of experience in computer science, evaluated the responses across the four categories. The labelling achieved a high inter-annotator agreement with a Cohen's kappa (K) score of 0.80, confirming the reliability of the evaluation framework.
These results evidence two critical areas where LLMs shine and their struggles, giving further information about their strengths and limitations concerning temporal reasoning.

\paragraph{Performance of Selected LLMs}
The evaluation of 12 language models, accessed through Hugging Face and OpenAI APIs, provided a comprehensive overview of their performance on temporal reasoning tasks. Small models like Llama-3.2-3B \cite{dubey2024llama3herdmodels} and Phi-3.5-mini \cite{abdin2024phi-3} gave bad performances, with 58\% and 66\% incorrect answers, respectively. Due to their restricted processing and resources, these models performed poorly in tokenization and reasoning. Mid-sized models, including Mistral-7B \cite{jiang2023mistral}, Llama-3-8B \cite{dubey2024llama3herdmodels}, and Llama-2-7B \cite{touvron2023llama}, demonstrated a more moderate improvement. They had trouble with complex reasoning problems, although they were able to improve their tokenization accuracy. Larger models, including Llama-3-70B \cite{dubey2024llama3herdmodels}, Qwen2.5-72B \cite{yang2024qwen2}, and Command R+ \cite{Cohere_Command_R_Plus}, were more robust in their performance, especially in date interpretation and logical reasoning. However, there were inconsistencies in specific formats. Proprietary models, including GPT-3.5 \cite{brown2020language}, GPT-4-turbo \cite{openai2023gpt-4}, GPT-4o, and GPT-4o-mini \cite{openai2024gpt-4o} outperformed all the rest, with GPT-4-turbo leading on correct responses with 63\% and the lowest rate of incorrect answers at 16\%. These results emphasise that model size, architecture, and diversity of pretraining data all bear on performance related to temporal reasoning tasks.

\paragraph{Performance Based on Date Formats}
The format of the date had a significant impact on model performance. Models performed best for formats that included clear separators and natural language cues, such as "YYYY, Mon DD" with 57\% correct and "DDMonYYYY" with 54\% correct. The poorest performance was from formats like "YYYY/DD (Julian)" and "DD/YYYY (Julian)", with only 31\% and 34\% correct, respectively, since the representation is less common and more complex in tokenization. This trend indicates format standardisation's apparent relevance in improving date processing efficiency in LLMs.

\paragraph{Performance Across Temporal Contexts}
Temporal context also mattered a lot. Models were better with future dates, 50\% correct, compared to historical dates, 44\%, and present dates, 35\%. This runs contrary to the expectations and may point to the fact that future-oriented reasoning tasks tap into the generative and predictive capabilities of the models. Historical and present contexts, which often require exact recall or conformity to training data, proved more difficult due to inconsistencies in the coverage of pretraining corpora.

\paragraph{Performance by Question Type}
Question type further modified results, with commonsense reasoning questions reaching the highest percentage of correctness: 51\%. These questions depended less on explicit tokenization and more on logical inference, which LLMs did comparatively well. Factual questions were at 45\%, while conceptual questions reached slightly lower performances of 40\%. Numerical reasoning questions were the hardest; only 37\% were correct since these often included some calculation or logical deduction that exposed the weaknesses in the models' reasoning capability.

\section{Discussion} \label{sec:discussion}


This study highlights the need for targeted strategies to address temporal biases in large language models (LLMs). A key step is to enhance pretraining datasets to ensure temporal diversity, incorporating historical, contemporary, and futuristic contexts. While resources like Redpajama \cite{weber2024redpajamaopendatasettraining} and Dolma \cite{soldaini2024dolmaopencorpustrillion} are open source, researchers should develop data focused on temporal reasoning with varied formats and cultural contexts.

Post-training methods, such as Direct Preference Optimization (DPO) \cite{rafailov2024directpreferenceoptimizationlanguage}, offer a promising avenue for fine-tuning models using curated datasets specifically designed to improve their logical temporal reasoning capabilities \cite{su2024timobettertemporalreasoning, tan2023benchmarkingimprovingtemporalreasoning}. These approaches can help align the models’ outputs with human-preferred logical reasoning patterns, addressing specific shortcomings in temporal tasks. Additionally, Retrieval-Augmented Generation (RAG) \cite{liu2024raghelpreasoningllm} enhances LLMs by integrating external knowledge dynamically during inference, allowing the models to access up-to-date or context-specific temporal information beyond their static training data. Moreover, prompting techniques such as Chain of Thought (CoT) prompting \cite{wei2023chainofthoughtpromptingelicitsreasoning} enable models to break down complex temporal reasoning tasks into incremental steps, improving interpretability and logical coherence \cite{liu2024raghelpreasoningllm, xiong2024largelanguagemodelslearn}.

However, while these post-training methods significantly mitigate biases in temporal reasoning and improve model performance, they are not sufficient to completely eliminate inherent biases. Factors such as the limitations of pre-trained embeddings, the static nature of foundational knowledge, and the variability in task-specific datasets mean that biases are likely to persist at some level. Thus, post-training approaches should be viewed as an important step toward reducing biases.


\section{Conclusion} \label{sec:conclusion}
Our paper addresses the challenges of temporal biases in large language models (LLMs) and proposes a structured approach to analyse their performance with temporal data. We introduced the DateLogicQA dataset and the Semantic Integrity Metric to evaluate the impact of diverse date formats and contexts on tokenization and reasoning. Our findings highlighted representation-level biases, where temporal contexts are inconsistently encoded, and logical-level biases, evident in varying outputs for similar prompts. We suggest mitigation strategies, such as temporally balanced pretraining datasets, post training and prompting methods.

\section*{Limitations} \label{sec:limitations}
\noindent\textbf{Future Scalability.}
The manual human evaluation approach for temporal reasoning performance analysis was time-consuming and challenging for future scalability. Furthermore, the evaluation technique requires high consensus among evaluators, especially when team size expands. Maintaining the evaluation quality in a larger team is also particularly difficult, and it might require more effort to cross-validate the results.

\section*{Ethical Considerations}

\noindent\textbf{AI usage.} It's pertinent to acknowledge the role of AI tools such as ChatGPT in our project. Specifically, Grammarly was utilized minimally and primarily for grammar corrections in our documents. This use was strictly confined to enhancing linguistic accuracy and improving the readability of our written materials. It's important to clarify that the core research, analysis, and development were conducted independently by our team.

\noindent\textbf{Human Annotation.} The human annotators involved in this project are professionals with expertise in computer science. No sensitive or personally identifiable data was used in the annotation process, adhering to ethical guidelines and data privacy standards. The human annotators are co authors on this paper.

\bibliography{acl2020,custom}

\appendix
\section{Appendix}

\begin{table*}[!htp]\centering
\scriptsize
\resizebox{\textwidth}{!}{%
\begin{tabular}{lrrrrrrrrrr}\toprule
\textbf{Format} &\textbf{Model} &\textbf{Date} &\textbf{Year} &\textbf{Time Period} &\textbf{Century} &\textbf{TC} &\textbf{Tokenized Output} &\textbf{SI} &\textbf{SC} &\textbf{PS} \\\midrule
MMDDYYYY &Baseline &10271606 &1606 &Historical (Pre-2000) &17th Century &3 &10 27 1606 &1.00 &false &true \\
MMDDYYYY &OLMoE &10271606 &1606 &Historical (Pre-2000) &17th Century &4 &10 27 16 06 &0.66 &true &true \\
MMDDYYYY &OLMo &10271606 &1606 &Historical (Pre-2000) &17th Century &4 &10 27 16 06 &0.66 &true &true \\
MMDDYYYY &Llama 3 &10271606 &1606 &Historical (Pre-2000) &17th Century &3 &102 716 06 &0.60 &true &true \\
MMDDYYYY &Llama 3.1 &10271606 &1606 &Historical (Pre-2000) &17th Century &3 &102 716 06 &0.60 &true &true \\
MMDDYYYY &Llama 3.2 &10271606 &1606 &Historical (Pre-2000) &17th Century &3 &102 716 06 &0.60 &true &true \\
MMDDYYYY &Davinci-003 &10271606 &1606 &Historical (Pre-2000) &17th Century &3 &1027 16 06 &0.60 &true &true \\
MMDDYYYY &GPT-3.5 &10271606 &1606 &Historical (Pre-2000) &17th Century &3 &102 716 06 &0.60 &true &true \\
MMDDYYYY &GPT-4o &10271606 &1606 &Historical (Pre-2000) &17th Century &3 &102 716 06 &0.60 &true &true \\
MMDDYYYY &GPT-4 &10271606 &1606 &Historical (Pre-2000) &17th Century &3 &102 716 06 &0.60 &true &true \\
MMDDYYYY &Cohere Aya &10271606 &1606 &Historical (Pre-2000) &17th Century &8 &1 0 2 7 1 6 0 6 &0.45 &true &true \\
MMDDYYYY &Gemma &10271606 &1606 &Historical (Pre-2000) &17th Century &8 &1 0 2 7 1 6 0 6 &0.45 &true &true \\
MMDDYYYY &DeepSeek &10271606 &1606 &Historical (Pre-2000) &17th Century &8 &1 0 2 7 1 6 0 6 &0.45 &true &true \\
MMDDYYYY &Cohere &10271606 &1606 &Historical (Pre-2000) &17th Century &8 &1 0 2 7 1 6 0 6 &0.45 &true &true \\
MMDDYYYY &Qwen &10271606 &1606 &Historical (Pre-2000) &17th Century &8 &1 0 2 7 1 6 0 6 &0.45 &true &true \\
MMDDYYYY &Phi 3.5 &10271606 &1606 &Historical (Pre-2000) &17th Century &9 &\_ 1 0 2 7 1 6 0 6 &0.40 &true &true \\
MMDDYYYY &Llama 2 &10271606 &1606 &Historical (Pre-2000) &17th Century &9 &\_ 1 0 2 7 1 6 0 6 &0.40 &true &true \\
MMDDYYYY &Mistral &10271606 &1606 &Historical (Pre-2000) &17th Century &9 &\_ 1 0 2 7 1 6 0 6 &0.40 &true &true \\
MMDDYYYY &Llama 1 &10271606 &1606 &Historical (Pre-2000) &17th Century &9 &\_ 1 0 2 7 1 6 0 6 &0.40 &true &true \\
\bottomrule
\end{tabular}}
\caption{Generated by Spread-LaTeX}\label{tab:alltokenized}
\end{table*}
\begin{figure}[t]
\centering
\includegraphics[width=1\columnwidth]{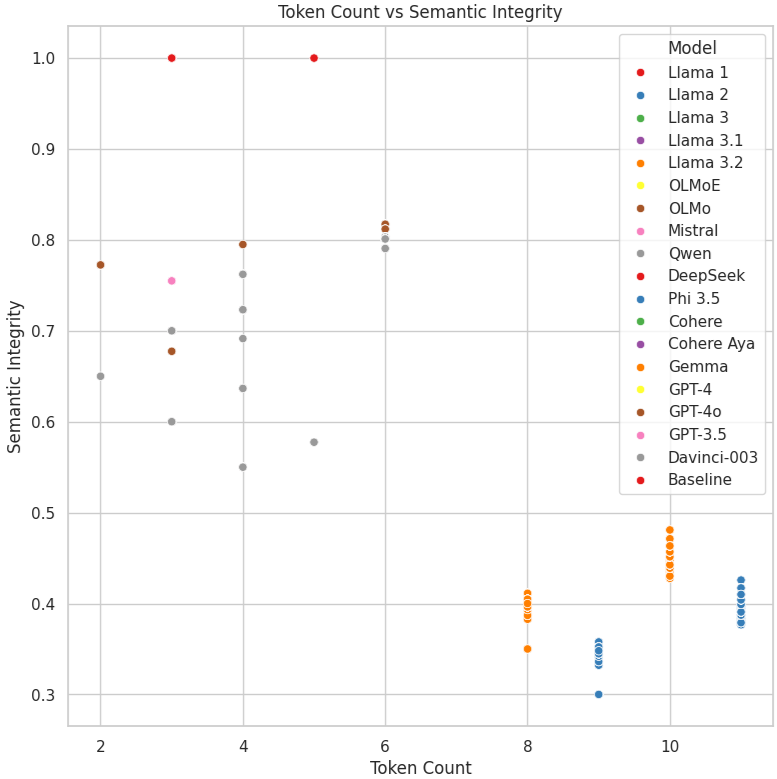}
\caption{Correlation plot between semantic integrity score against token count}
\label{fig:SemanticTokenCount}
\end{figure}

\end{document}